# Image Processing Based Systems and Techniques for the Recognition of Ancient and Modern Coins

Shatrughan Modi
Frontier Research Group, Samsung India Software Operations,
Bangalore- 560093, India

Seema Bawa, PhD.
Dept. of Computer Science and Engineering
Thapar University
Patiala-147004, India.

## ABSTRACT
Coins are frequently used in everyday life at various places like in banks, grocery stores, supermarkets, automated weighing machines, vending machines etc. So, there is a basic need to automate the counting and sorting of coins. For this machines need to recognize the coins very fast and accurately, as further transaction processing depends on this recognition. Three types of systems are available in the market: Mechanical method based systems, Electromagnetic method based systems and Image processing based systems. This paper presents an overview of available systems and techniques based on image processing to recognize ancient and modern coins.

## 1. INTRODUCTION
There is a basic need of highly accurate and efficient automatic coin recognition systems in our daily life. Coin recognition systems and coin sorting machines have become a vital part of our life. They are used in banks, supermarkets, grocery stores, vending machines etc. In-spite of daily uses coin recognition systems can also be used for the research purpose by the institutes or organizations that deal with the ancient coins. There are three types of coin recognition systems based on different methods used by them available in the market:

- Mechanical method based systems
- Electromagnetic method based systems
- Image processing based systems

The mechanical method based systems use parameters like diameter or radius, thickness, weight and magnetism of the coin to differentiate between the coins. But these parameters can not be used to differentiate between the different materials of the coins. It means if we provide two coins one original and other fake having same diameter, thickness, weight and magnetism but with different materials to mechanical method based coin recognition system then it will treat both the coins as original coin so these systems can be fooled easily.

The electromagnetic method based systems can differentiate between different materials because in these systems the coins are passed through an oscillating coil at a certain frequency and different materials bring different changes in the amplitude and direction of frequency. So these changes and the other parameters like diameter, thickness, weight and magnetism can be used to differentiate between coins. The electromagnetic based coin recognition systems improve the accuracy of recognition but still they can be fooled by some game coins.

In the recent years coin recognition systems based on images have also come into picture. In these systems first of all the image of the coin to be recognized is taken either by camera or by some scanning. Then these images are processed by using various techniques of image processing like FFT, DCT, edge detection, segmentation etc. and further various features are extracted from the images. Based on these features different coins are recognized. This paper presents existing systems and techniques proposed by various researchers on image based coin recognition.

## 2. APPROACHES / TECHNIQUES
There are various approaches proposed by various researchers for image based coin recognition. We can broadly classify these approaches based on the coins (ancient coins, modern coins, ancient and modern both) on which they can be applied.

### 2.1 Approaches for Modern Coins
Most of the approaches proposed till now can be applied for recognition of modern coins.

*In 1992 [1] Minoru Fukumi et al.* presented a rotational invariant neural pattern recognition system for coin recognition. They have used 500 yen coin and 500 won coin to perform the experiment. In this work they have created a multilayered neural network and a preprocessor consists of many slabs of neurons. This preprocessor was used to get a rotational invariant input for the multilayered neural network. For the weights of neurons in preprocessor, concept of circular array was used instead of square array. The results show that 25 slabs with 72 neurons in each slab give the best recognition.

*In 1993 [2] Minoru Fukumi et al.* tried to achieve 100% accuracy for coins. They have used 500 yen coin and 500 won coin. In this work they have used Back Propagation (BP) and Genetic Algorithm (GA) to design neural network for coin recognition. BP is used to train the network. Then after training, GA is used to reduce the size of network by varying the architecture to achieve 100% recognition accuracy rate.

*Paul Davidsson [3] in 1996* presented an approach for coin classification using learning characteristic decision trees by controlling the degree of generalization. Decision trees constructed by ID3-like algorithms were unable to detect instances of categories not present in the set of training examples. Instead of being rejected, such instances get assigned to one of the classes actually present in the training set. To solve this problem the algorithm with learning characteristic, rather than discriminative, category descriptions was proposed. In addition, the ability to control the degree of generalization was identified as an essential property of such algorithms. Experiments were performed on Canadian and Hong-Kong coins and accuracy of 99.7% for Canadian and 98.3% for Hong-Kong coins was achieved.

*Michael Nolle et al. [4]* at the ARC Seibersdorf research centre in 2003 developed a coin recognition and sorting system called Dagobert. This system was designed for fast classification of large number of modern coins from 30 different countries. Coin classification was accomplished by correlating the edge image of the coin with a pre-selected subset of master coins and finding the master coin with lowest distance. Pre-selection of master coins was done based on three rotation-invariant features (edge angle distribution, edge distance distribution, occurrences of different rotation-invariant patterns on circles centered at edge pixels), coin diameter and thickness. Experiments on 12,949 coins were performed and 99.24% recognition rate was achieved.

A coin recognition system to recognize US coins using vector quantization and histogram modeling was presented by *Seth McNeill et al. [5] in 2004*. The system mainly focuses on the texture of various images imprinted on the coin tail. Based on different image texture the system differentiate between Bald eagle on the quarter, the Torch of liberty on the dime, Thomas Jefferson's house on the nickel, and the Lincoln Memorial on the penny. Experiments show that out of 200 coin images 188 were correctly classified. Thus, 94% recognition accuracy rate was achieved.

In 2005 a multistage approach for coin classification using Eigenspace and Bayesian fusion was presented by *Reinhold Huber et al. [6]*. In the first stage, a translational and rotational invariant description is computed. In a second stage, an illumination-invariant eigenspace is selected and probabilities for coin classes are derived for both sides of each coin. In the final stage, coin class probabilities for both coin sides are combined through Bayesian fusion including a rejection mechanism. Experiments show that 93.23% of 11,949 coins from thirty different countries were correctly classified.

*In 2005 [7] R. Bremananth et al.* presented an approach using neural network for coin recognition. In this work they have concentrated on the recognition of the numerals on the coin rather than other images. For this they extract a sub image of numeral from coin then this sub image is used for character recognition. To achieve rotation invariance Gabor filters and Back Propagation neural network are used. Experiments are performed on 1-rupee, 2-rupee and 5-rupee coin. The experiments show 92.43% recognition accuracy rate.

*L.J.P. van der Maaten et al. [8] in 2006* developed a fast system for reliable coin classification called COIN-O-MATIC. In this system coin classification is done based on edge-based statistical features (edge angle distributions, edge distance distributions, edge angle-distance distributions). The system consists of four subsystems: (1) a segmentation subsystem, (2) a feature extraction subsystem, (3) a classification subsystem, and (4) a verification subsystem. Experiments were performed on MUSCLE-CIS dataset and 72% classification accuracy was achieved.

*Adnan Khashman et al. [9, 10]* presented an Intelligent Coin Identification System (ICIS) in 2006. ICIS uses neural network and pattern averaging for recognizing rotated coins at various degrees. ICIS consists of two phases. First is image processing in which coin images are mode converted, cropped, compressed, trimmed, pattern averaged etc. This is the preprocessing phase. In second phase a back propagation neural network is trained. Once neural network converges and learns then only one forward pass is used that yields the identification results. ICIS shows very encouraging results. It shows 96.3% correct identification i.e. 77 out of 80 variably rotated coin images were correctly identified. ICIS is very effective in reducing time costs because it reduces the processing data by preprocessing images.

*In 2006 [12] P.Thumwarin et al.* presented a robust method for coin recognition with rotation invariance. In this method the rotation invariance feature is represented by the absolute value of Fourier coefficients of polar image of coin on circles with different radii. In this paper the variations on surface of coin such as light reflection effect are also considered. These effects can be reduced by Fourier approximation of the coin image. Finally coins are recognized by calculating distance between the absolute value of Fourier coefficients obtained from the reference coin and the coin to be recognized.

A fast and reliable coin recognition system based on registration approach was presented by *Marco Reisert et al. [13]* in 2007. For this gradient directions are used whereas gradient magnitude is completely neglected. Then classification is performed by simple nearest neighbor classifier scheme followed by several rejection criteria to meet the demand of low false positive rate. The system presented was also reliable to illumination and contrast changes. Experiments were performed on CIS benchmark dataset.

*In 2009 [18] Linlin Shen et al.* presented an image based approach for coin recognition. In this paper Gabor wavelets are used to extract features. Coin image is divided into number of small sections by concentric ring structures. Then Gabor coefficients from each section are concatenated to form feature vectors. Then these feature vectors are compared with the feature vectors of saved images in database. Identification is done using Euclidean distance and the nearest neighbor classifier. For experiments public MUSCLE database consisting of over 10,000 images is used. This algorithm shows the recognition accuracy rate of 74.27%.

*CHEN Cai-ming et al. [19] in 2010* presented a coin recognition system with rotation invariance. In this system same approach presented in [12] is used but with BP neural network i.e. the feature vector which is the input to BP neural network is obtained by calculating the absolute value of Fourier coefficients of polar image of coin on circles with different radii. The experiments show the accuracy rate of 83%.

An efficient method for coin recognition using a statistical approach was presented in 2010 by *Hussein R. Al- Zoubi* [20]. In this paper statistical methods are used to recognize Jordanian coins. In this approach main focus is on area and color of coins. First of all coin image is converted to gray level image then the gray image is segmented into two regions coin and background based on the histogram of image. Then segmented image is cleaned and then four parameters i.e. area, average red, average green and average blue are calculated. Then based on these parameters decision is made that to which category the coin belongs. This approach yields high recognition rate of 97% which is very encouraging.

*In 2010 [21] Huahua Chen* presented an approach for Chinese coin recognition based on unwrapped image and rotation invariant template matching. In this approach first of all coin segmentation is done using Hough transform then the segmented image is unwrapped. Unwrapping is done by transforming reference and specimen coin image from Cartesian coordinates to polar coordinates. After unwrapping, the template matching is done and on the basis of this recognition is done. Experiments were performed on 144 variably rotated coins images. Out of which 116 were correctly recognized. So, overall 80.6% correct recognition was achieved.

*In 2011 [22] Vaibhav Gupta et al.* presented an approach based on image subtraction technique for recognition of Indian coins. In this approach system performs 3 checks (radius, coarse and fine) on the input coin image. First of all radius is calculated of the input image. Then based on the radius a test image (rotated at certain fixed angle) from database is selected. Then coarse image subtraction between object and test image is done. Then, minima of the resultant image is checked if it is less than a specified threshold then fine image subtraction between object and test image is done otherwise new test image is selected. Then based on fine image subtraction, recognition takes place.

## 2.2 Approaches for Ancient Coins

There is very less work done on recognition of ancient coins. The main reason for this is that the ancient coins do not have symmetrical boundaries like modern coins because ancient coins were hammered or casted during manufacturing whereas modern coins are minted. Also ancient coins are generally found in poor conditions due to wear or fouling. So due to irregular shape and poor condition, the general approaches of coin recognition easily fail for ancient coins.

*Kaiping Wei et al. [14] in 2007* presented a novel approach for classification of ancient coins based on image textual information. For extracting textual information Tree-Structured Wavelet Transform (TWT) and Ant Colony Optimization (ACO) algorithm is used. The multi-resolution character of the texture is extracted by TWT, and information can be accessed in various scale rather than low frequency. In addition, segmentation algorithm based on ACO is implemented before TWT to obtain textural information with the absence of noise. The results show that this hybrid approach provides very accurate recognition results for ancient coins.

## 2.3 Approaches for both (Modern and Ancient) Coins

There are some approaches that can be applied to both ancient and modern coins.

*In 2006 [11] Laurens J.P. van der Maaten et al.* presented algorithms for automatic coin classification. The algorithms take digital images of coins as input and generate a class as output i.e. to which class the coin belongs. There are two stages of automatic classification. First is feature extraction stage and second is classification stage. In first stage i.e. in feature extraction stage two types of features are extracted. First are Contour features and second are Texture features. For extracting Contour features first of all contour image is extracted from original image of coin and then this contour image is represented in statistical features using multi-scale edge angle histograms and multi-scale edge distance histograms. For extracting texture features two types of wavelet features i.e. Gabor wavelet features and Daubechies wavelet features are used. For experiments two datasets are used in this paper. The main dataset is the MUSCLE CIS dataset, which is used for evaluating the effectiveness of the feature types. The second dataset is the Merovingen coin dataset which is employed to evaluate to what extent our feature types are appropriate for ancient coin classification. The results revealed that a combination of Contour and Texture features yield the best performance.

*Abdolah Chalechale [15]* presented a novel approach for coin image recognition using image abstraction and spiral decomposition in 2007. The approach SDAI (Spiral Decomposition of Abstract Image) enables measuring the similarity between full color multi- component coin images and need no cost intensive image segmentation. Here an abstract image is derived from original image based on strong edges of the coin. Then spiral distribution of pixels in the abstract image is employed as the key concept for feature extraction. Extracted features are scale, translation and rotation invariant. The images used for query set and test database are scanned, photographed or collected from web. The proposed approach is compared with three other approaches i.e. QVE, PFD (Polar Fourier Descriptor) and EHD (Edge Histogram Distribution). The results show that the proposed approach is much better than other three approaches because it shows significant improvement in recall ratio using proposed features.

*In 2007 [16, 17] Martin Kampel et al.* gives the overview and the preliminary results of EU project COINS (COmbatting Illicit Numismatic Sales). The project aims to substantially contribute to the fight against illegal trade and theft of coins which appears to be a major part of the illegal antiques market. In this paper recognition of both ancient and modern coins is considered. They have also discussed the existing approaches like Eigenspace approach, Contour based algorithms and Gradient based algorithms. They use two approaches for segmentation: Edge based segmentation and Generalized Hough Transform (GHT). Then they use edge based statistical distribution to extract the features. Then they just compare the features to recognize the coin by using K-nearest Neighbor algorithm. They have also compared the results from both Edge based segmentation and GHT and clearly the former performed better than the later. Table 1 presents a chronological summary of techniques developed for automatic recognition of ancient as well as modern coins.

## 3. CONCLUSION

This paper presents various systems developed and existing techniques for coin recognition based on image processing. A comparison of these techniques has been given in Table 1. There are so many techniques proposed till now for modern coins, and maximum accuracy of 99.7% has been achieved in 1996 for Canadian coins using decision trees, but still very less work has been done for the recognition of ancient coins.

**Table 1. Chronological Development of Techniques for Modern and / or Ancient Coins**

| Sr. No. | Year | Technique Used | Dataset of Coins Used | Modern/Ancient Coins | | Accuracy Achieved |
|---|---|---|---|---|---|---|
| | | | | Modern | Ancient | |
| 1 | 1992 [1] | Neural network, circular array | 500 yen (Japan) and 500 won (Korea) coin | √ | | |
| 2 | 1993 [2] | Neural network using Genetic Algorithm | 500 yen (Japan) and 500 won (Korea) coin | √ | | |
| 3 | 1996 [3] | Decision trees | Canadian and Hong Kong coins | √ | | 99.7% - Canadian coins, 98.3% - Hong Kong coins |
| 4 | 2003 [4] | Edge angle distribution, edge distance distribution | Coins from 30 countries | √ | | 99.24% |
| 5 | 2004 [5] | Vector Quantization and Histogram Modeling | US coins | √ | | 94% |
| 6 | 2005 [6] | Eigenspaces and Bayesian fusion | Coins from 30 countries | √ | | 93.23% |
| 7 | 2005 [7] | Neural network, Gabor filter, Statistical color threshold | Indian coins | √ | | 92.43% |
| 8 | 2006 [8] | Edge angle distribution, edge distance distribution, edge angle-distance distribution | MUSCLE CIS dataset | √ | | 72% |
| 9 | 2006 [9, 10] | Neural Network, Pattern Averaging | Turkish 1 Lira and 2 Euro coin | √ | | 96.3% |
| 10 | 2006 [11] | Multi-scale edge angle histograms, Multi-scale edge distance histograms, Gabor wavelet, Daubechies wavelet | MUSCLE CIS dataset and Merovingen coin dataset | √ | √ | 76% |
| 11 | 2006 [12] | Fourier approximation of polar image | Thai amulet and Thai baht coins | √ | | |
| 12 | 2007 [14] | Tree structured wavelet transform, Ant colony optimization algorithm | | | √ | |
| 13 | 2007 [13] | Registration approach based on gradient directions | CIS Benchmark dataset | √ | | |
| 14 | 2007 [15] | Image abstraction and spiral decomposition | COIN BANK | √ | √ | |
| 15 | 2007 [16, 17] | Edge based segmentation, Generalized hough transform and K-nearest neighbor algorithm | MUSCLE CIS dataset | √ | √ | 76% |
| 16 | 2009 [18] | Gabor wavelet, Euclidean distance and nearest neighbor classifier | MUSCLE dataset | √ | | 74.27% |
| 17 | 2010 [19] | Fourier approximation of polar image, neural network | Chinese coins | √ | | 83% |
| 18 | 2010 [20] | Statistical approach | Jordanian coins | √ | | 97% |
| 19 | 2010 [21] | Rotation invariant template matching | Chinese coins | √ | | 80.6% |
| 20 | 2011 [22] | Image subtraction technique | Indian coins | √ | | |

## 4. REFERENCES

[1] Fukumi M. and Omatu S., "Rotation-Invariant Neural Pattern Recognition System with Application to Coin Recognition", IEEE Trans. Neural Networks, Vol.3, No. 2, pp. 272-279, March, 1992

[2] Fukumi M. and Omatu S., "Designing A Neural Network For Coin Recognition By A Genetic Algorithm", Proceedings of 1993 International Joint Conference on Neural Networks, 1993:2109-2112.

[3] P. Davidsson, "Coin classification using a novel technique for learning characteristic decision trees by controlling the degree of generalization", Ninth International Conference on Industrial & Engineering Applications of Artificial Intelligence & Expert Systems, 1996.

[4] M. Nolle, H. Penz, M. Rubik, K. Mayer, I. Hollander and R. Granec, "Dagobert – A New Coin Recognition and Sorting System", Proceedings of the 7th International Conference on Digital Image Computing - Techniques and Applications (DICTA'03), Syndney, Australia.

[5] McNeill S., Schipper J., Sellers T. and Nechyba M.C.. "Coin Recognition using Vector Quantization and Histogram Modeling". 2004 Florida Conference on Recent Advances in Robotics (FCRAR)

[6] Reinhold Huber, Herbert Ramoser, Konrad Mayer, Harald Penz, Michael Rubik, "Classification of coins using an eigenspace approach", Pattern Recognition Letters (2005), Vol.26, No.1, 61-75

[7] R. Bremananth, B. Balaji, M. Sankari and A. Chitra, "A new approach to coin recognition using neural pattern analysis" IEEE Indicon 2005 Conference, Chennai, India, 11 - 13 Dec. 2005.

[8] L.J.P. van der Maaten, P.J. Boon, "COIN-O-MATIC: A fast system for reliable coin classification", MUSCLE CIS Coin Recognition Competition Workshop 2006

[9] Khashman A., Sekeroglu B. and Dimililer K., "Intelligent Coin Identification System", Proceedings of the IEEE International Symposium on Intelligent Control ( ISIC'06 ), Munich, Germany, 4-6 October 2006.

[10] Khashman A., Sekeroglu B. and Dimililer K., "ICIS: A Novel Coin Identification System", Lecture Notes in Control and Information Sciences, Vol. 345, Springer-Verlag, September 2006.

[11] L.J.P. van der Maaten and E.O. Postma, "Towards automatic coin classification", Proceedings of the EVA-Vienna 2006, Vienna, Austria, 2006.

[12] Thumwarin, P., Malila, S., Janthawong, P. and Pibulwej, W., "A Robust Coin Recognition Method with Rotation Invariance", 2006 International Conference on Communications, Circuits and Systems Proceedings, 2006.

[13] Reisert M., Ronneberger O. and Burkhardt H., "A Fast and Reliable Coin Recognition System", in Proceedings of the 29th DAGM conference on Pattern recognition, 2007

[14] Kaiping Wei, Bin He, Fang Wang, Tao Zhang and Quanjun Ding, "A Novel Method for Classification of Ancient Coins Based on Image Textures," dmamh, pp.63-66, Second Workshop on Digital Media and its Application in Museum & Heritage (DMAMH 2007), 2007

[15] Chalechale, A. "Coin Recognition Using Image Abstraction And Spiral Decomposition", 9th International Symposium on Signal Processing and Its Applications, 2007. ISSPA 2007.

[16] Zaharieva, M., Kampel, M. and Zambanini, S., "Image based recognition of coins – An Overview of the COINS project.", 31st AAPR/OAGM Workshop, Krumbach, Austria, 2007.

[17] Kampel, M. and Zambanini, S., "Optical Recognition of Modern and Roman Coins", Layers of Perception- CAA 2007.

[18] Shen, L., Jia, S., Ji, Z. and Chen, W.S., "Statistics of Gabor features for coin recognition", IEEE International Workshop on Imaging Systems and Techniques, 2009.

[19] Cai-ming Chen, Shi-qing Zhang, Yue-fen Chen, "A Coin Recognition System with Rotation Invariance," 2010 International Conference on Machine Vision and Human-machine Interface, 2010

[20] Al-Zoubi, H.R., "Efficient coin recognition using a statistical approach", 2010 IEEE International Conference on Electro/Information Technology (EIT), 2010.

[21] Chen, H. "Chinese Coin Recognition Based on Unwrapped Image and Rotation Invariant Template Matching", Third International Conference on Intelligent Networks and Intelligent Systems, 2010.

[22] Gupta, V., Puri, R., Verma, M., "Prompt Indian Coin Recognition with Rotation Invariance using Image Subtraction Technique", International Conference on Devices and Communications (ICDeCom), 2011.